\documentclass[letterpaper,11pt]{article}

\usepackage[margin=1in,includeheadfoot]{geometry}
\usepackage[utf8]{inputenc}
\usepackage[T1]{fontenc}    % use 8-bit T1 fonts
\usepackage[sorting=none]{biblatex}  % 使用 biblatex
\usepackage{hyperref}       % hyperlinks
\usepackage{url}            % simple URL typesetting
\usepackage{booktabs}       % professional-quality tables
\usepackage{tabularx}
\usepackage{makecell}
\usepackage{amsfonts}       % blackboard math symbols
\usepackage{nicefrac}       % compact symbols for 1/2, etc.
\usepackage{microtype}      % microtypography
\usepackage{lipsum}
\usepackage{graphicx}
\usepackage{amsmath}
\graphicspath{ {./images/} }

\title{Breaking Thought Patterns: A Multi-Dimensional Reasoning Framework for LLMs}
\author{Xintong~Tang, % 注意 ~ 可防止换行
  Meiru~Zhang,
  Shang~Xiao,
  Junzhao~Jin,\\   % 换行把作者组和单位分开
  Zihan~Zhao,
  Liwei~Li,
  Yang~Zheng,
  Bangyi~Wu\\[4pt]
  Behavision, China
  }

\begin{document}
    \maketitle	
    \begin{abstract}
    Large language models (LLMs) are often constrained by rigid reasoning processes, limiting their ability to generate creative and diverse responses. To address this, a novel framework called LADDER is proposed, combining Chain-of-Thought (CoT) reasoning, Mixture of Experts (MoE) models, and multi-dimensional up/down-sampling strategies which breaks the limitations of traditional LLMs. First, CoT reasoning guides the model through multi-step logical reasoning, expanding the semantic space and breaking the rigidity of thought. Next, MoE distributes the reasoning tasks across multiple expert modules, each focusing on specific sub-tasks. Finally, dimensionality reduction maps the reasoning outputs back to a lower-dimensional semantic space, yielding more precise and creative responses.
    Extensive experiments across multiple tasks demonstrate that LADDER significantly improves task completion, creativity, and fluency, generating innovative and coherent responses that outperform traditional models. Ablation studies reveal the critical roles of CoT and MoE in enhancing reasoning abilities and creative output. This work contributes to the development of more flexible and creative LLMs, capable of addressing complex and novel tasks.
    \end{abstract}

    \begin{figure}[htbp]            % h=here  t=top  b=bottom  p=page of floats
    \centering                    % 图片水平居中
    \includegraphics[width=\linewidth]{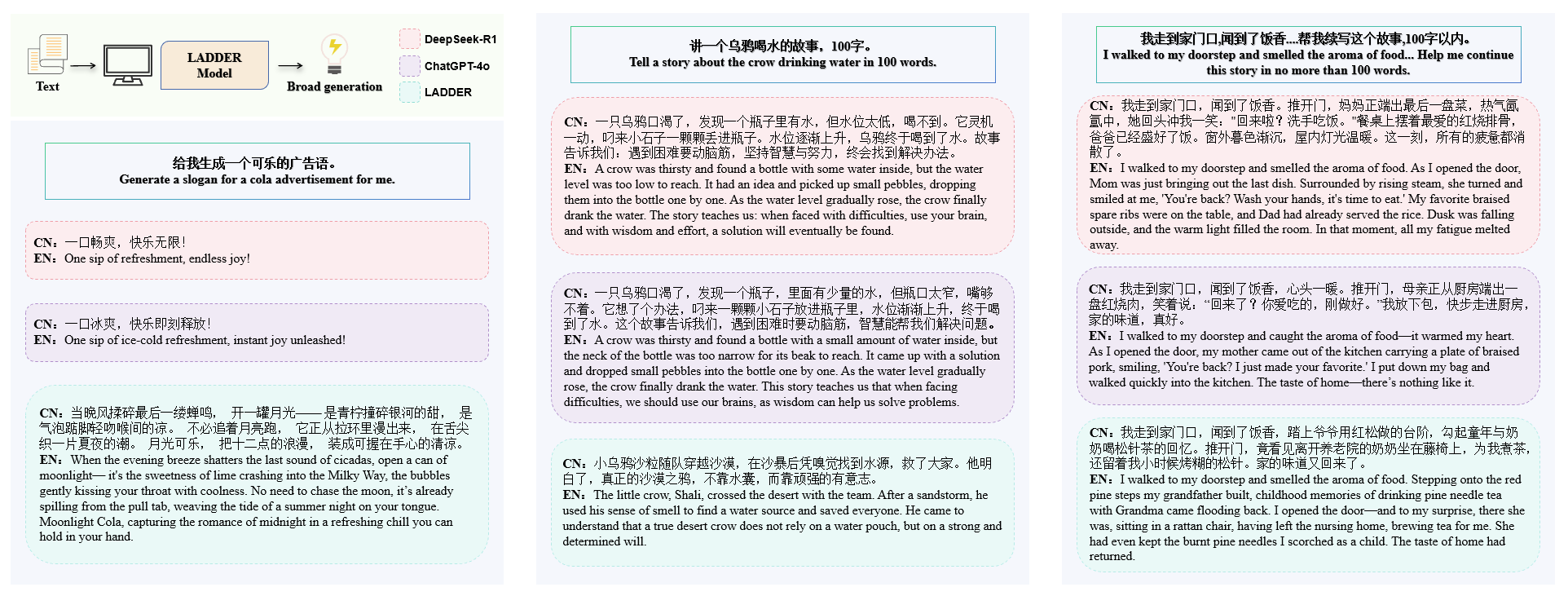}
    \caption{Comparative performance of large language models (LLMs) on advertising-copy and narrative-generation tasks: an analysis of the outputs produced by DeepSeek-R1 (pink), ChatGPT-4o (purple), and LADDER (cyan) in response to the prompts “cola advertisement,” “The Crow and the Pitcher” story, and “story continuation: smelling the aroma of food at the doorway.”}      % 图题
    \label{fig:1}             % 交叉引用标签
    \end{figure}
	
    \section{Introduction}
    With the rapid advancement of deep learning, the field of Natural Language Processing (NLP) has witnessed significant breakthroughs in tasks such as text generation and semantic understanding. Particularly, Transformer-based language models \cite{10.5555/3295222.3295349,devlin-etal-2019-bert} have greatly accelerated performance improvements across a wide range of applications. However, despite the impressive capabilities of large-scale pre-trained models such as GPT-4 \cite{openai2024gpt4technicalreport} and LLaMA-3 \cite{grattafiori2024llama3herdmodels} in language generation and pattern recognition, they continue to face critical challenges in tasks requiring creative reasoning and deep semantic control. These models typically function as black boxes, lacking transparent reasoning processes, which limits their applicability in scenarios that demand interpretability and the construction of causal chains \cite{geva2021transformerfeedforwardlayerskeyvalue}. Moreover, they often exhibit semantic fragility when handling advanced linguistic phenomena such as metaphor and analogy, and struggle to maintain semantic consistency in cross-domain knowledge transfer settings \cite{stevenson2025largelanguagemodelsgeneralize}. The monolithic Transformer architecture further lacks the capacity for modular knowledge representation and dynamic reasoning path selection, rendering it inflexible in adapting to task complexity \cite{stevenson2025largelanguagemodelsgeneralize,rae2022scalinglanguagemodelsmethods}.
    \\These limitations become especially pronounced in open-domain story generation. Empirical studies have demonstrated that state-of-the-art language models often rely on repetitive narrative structures when generating long-form texts, exhibiting signs of cognitive rigidity \cite{geva2021transformerfeedforwardlayerskeyvalue}. For instance, analysis of creative writing samples generated by GPT-4 and DeepSeek-R1 \cite{deepseekai2025deepseekr1incentivizingreasoningcapability} reveals a high degree of narrative template repetition, indicating a tendency toward path dependency during generation \cite{stevenson2025largelanguagemodelsgeneralize}. This dependency manifests not only in rigid plot progression but also in the limited flexibility of character relationship modeling and uniformity in narrative style control. Owing to the lack of systematic planning capabilities and dynamic mechanisms for operating within the semantic space, existing models struggle to adapt narrative strategies contextually, resulting in outputs that fall short in both creativity and adaptability \cite{stevenson2025largelanguagemodelsgeneralize,rae2022scalinglanguagemodelsmethods}.
    \\To address these key limitations in complex reasoning and generative tasks, this study proposes LADDER—a novel framework that integrates multi-level reasoning mechanisms with dynamic expert collaboration. This architecture is the first to deeply combine Chain-of-Thought (CoT) reasoning \cite{wei2023chainofthoughtpromptingelicitsreasoning} with the Mixture of Experts (MoE) paradigm \cite{shazeer2017outrageouslylargeneuralnetworks}, utilizing a modular design to decompose complex language tasks into interpretable cognitive stages, each executed by specialized subnetworks. The system initiates by leveraging the CoT mechanism to perform in-depth input analysis and task decomposition, explicitly delineating reasoning paths and intermediate goals. It then employs an attention-based multi-granularity gating network to dynamically activate the most relevant expert modules, enabling fine-grained knowledge routing. Finally, an adaptive semantic scaling mechanism is constructed to achieve continuous mapping from abstract concepts to concrete expressions, thereby supporting a bidirectional process of semantic abstraction and output anchoring.
    \\This research offers three primary contributions:
    \begin{itemize}
    \item Methodological Innovation. It presents the first deep integration of CoT reasoning and MoE architectures, significantly enhancing model interpretability and task adaptability \cite{wei2023chainofthoughtpromptingelicitsreasoning,shazeer2017outrageouslylargeneuralnetworks}.
    \item Technical Advancement. It introduces a novel multi-granularity gating network based on attention mechanisms, improving the selection efficiency and coordination among expert modules.
    \item Practical Impact. Systematic experimental evaluation demonstrates the framework’s superior performance across a range of NLP tasks, particularly showcasing strong transferability in few-shot learning scenarios.
    \end{itemize}
    Experimental results demonstrate that LADDER exhibits outstanding performance across various natural language processing tasks. In story generation tasks, the system produces texts that maintain semantic coherence while demonstrating greater creativity and logical rigor. Quantitative analysis shows that, compared to traditional methods, the proposed framework achieves significant improvements in text quality evaluation metrics and exhibits stronger adaptability in few-shot learning scenarios. These findings not only validate the effectiveness of the proposed framework but also offer new insights into the interpretability and adaptability of NLP systems.

    \section{Related Work}
    
    \subsection{Chain-of-Thought Reasoning}
    Reasoning enhancement techniques have made significant progress in improving the capabilities of large language models (LLMs) in recent years. Among them, CoT prompting was first proposed to enhance LLM performance on arithmetic tasks, markedly increasing accuracy on complex reasoning problems \cite{wei2023chainofthoughtpromptingelicitsreasoning}. This method explicitly generates intermediate reasoning steps, effectively activating the model’s step-by-step inference ability. Building upon this, researchers have developed various optimizations around the CoT strategy: Zhou et al. introduced automated prompt generation methods to reduce manual design costs \cite{zhou2023leasttomostpromptingenablescomplex}; Kojima et al. proposed a zero-shot CoT approach that triggers reasoning capabilities with only concise prompts \cite{kojima2023largelanguagemodelszeroshot}; Yao et al. further advanced the framework by introducing Tree-of-Thought, which extends linear reasoning paths into branched structures, enhancing flexibility and complexity control in reasoning \cite{yao2023treethoughtsdeliberateproblem}.
    Moreover, with the development of multimodal modeling techniques, CoT methods have been extended to vision-language joint reasoning scenarios, demonstrating potential for cross-modal cognition \cite{zhang2024multimodalchainofthoughtreasoninglanguage}. However, despite preliminary successes in areas such as mathematics and commonsense reasoning, current CoT-based research remains relatively limited in creative text generation and multi-stage content planning tasks, indicating a pressing need to further explore its application boundaries and scalability in complex generative tasks.
    
    \subsection{Mixture of Experts}
    Modular architectures have recently emerged as a key direction for enhancing the performance of large language models. Among them, MoE models show great potential in optimizing computational efficiency through dynamic activation of specialized sub-networks \cite{shazeer2017outrageouslylargeneuralnetworks}. Early work by Shazeer et al. introduced a sparse gating mechanism that laid the foundation for training large-scale MoE models \cite{shazeer2017outrageouslylargeneuralnetworks}. Subsequently, Zhou et al. proposed an expert routing strategy that effectively alleviated load imbalance, improving both routing robustness and resource utilization \cite{zhou2022mixtureofexpertsexpertchoicerouting}. In open-source implementations, the Mixtral model developed by Jiang et al. demonstrated the scalability and deployment potential of efficient MoE architectures in large language models \cite{jiang2024mixtralexperts}.
    Although MoE systems have achieved significant progress in computation and scalability, mainstream research still primarily focuses on routing mechanism optimization, training stability, and inference efficiency. In contrast, relatively little attention has been paid to semantic collaboration among experts and their potential for innovation in reasoning. The LADDER framework aims to fill this gap by exploring the collaborative reasoning capabilities of expert modules in complex cognitive tasks, thereby promoting an architectural shift from efficiency-oriented to creativity-oriented design.
    
    \subsection{Dimensionality Techniques}
    In the field of Natural Language Processing (NLP), semantic space transformation techniques have evolved from early linear dimensionality reduction approaches to more sophisticated and semantically-aware representation strategies. Early studies, such as the PCA-based dimensionality reduction method proposed by Raunak, demonstrated that vector compression could be achieved without significant performance degradation, making it a classical solution for embedding optimization \cite{raunak2017simpleeffectivedimensionalityreduction}. With the increasing demand for expressive representations, researchers began to explore semantic dimensionality expansion. For example, Ma and Cambria proposed a concept embedding approach that projects words into a conceptual space to enhance higher-order semantic understanding \cite{ma2018conceptbasedembeddingsnaturallanguage}. Hasan and Curry introduced a manifold-preserving dimensionality reduction strategy, which improves the quality of word embeddings while retaining the local topological structure of the representation space \cite{hasan-curry-2017-word}. More recently, Mu et al. proposed a weakly supervised semantic projection method, which enables dynamic dimensional adjustment under controllable performance loss, thus enhancing model adaptability in low-resource scenarios \cite{mu2024learningcompresspromptsgist}.
    Building upon these developments, the LADDER framework introduces a novel integration of semantic expansion and compression with its reasoning modules, enabling dynamic switching and efficient representation of semantic spaces across different cognitive stages. This design constructs a more adaptable and interpretable path for language generation.

    \subsection{Emergent Behavior in LLMs}
    The emergent behavior exhibited by Large Language Models (LLMs) as their training scale significantly increases has become a critical frontier in understanding their capability boundaries and reasoning potential. Emergence refers to the phenomenon where a model begins to exhibit qualitatively new abilities or properties once its parameter count, training data size, or computational complexity surpasses a certain threshold \cite{Ganguli_2022}. This concept was first systematically defined by Ganguli et al. (2022) while analyzing performance phase transitions in language models, and later empirically validated by Wei et al. (2022) through multi-task benchmark evaluations. For instance, GPT-3 demonstrated a sudden surge in capabilities—such as multi-step mathematical reasoning, few-shot learning, and code generation—after exceeding $2 \times 10^{22}$ FLOPs during training \cite{wei2022emergentabilitieslargelanguage, brown2020languagemodelsfewshotlearners}. These capabilities do not scale linearly with model size, but rather emerge abruptly, showing characteristics similar to phase transitions.
    In terms of underlying mechanisms, Power et al. (2023) argued that certain emergent capabilities are not merely the result of shallow pattern memorization, but stem from the model’s construction of more complex symbolic transformation structures. For example, in modular arithmetic tasks, the model may implicitly learn advanced mathematical tools such as Fourier transforms to achieve generalization \cite{power2022grokkinggeneralizationoverfittingsmall}.
    However, the emergence of such capabilities is also accompanied by significant risks. Bommasani et al. (2022) pointed out that as model scale increases, so too do risks associated with output bias and hallucinations—issues that are difficult to eliminate through conventional fine-tuning methods \cite{bommasani2022opportunitiesrisksfoundationmodels}. Therefore, enhancing the interpretability and predictability of emergent behavior has become a central challenge in LLM safety research.
    To improve controllability over emergence, current research mainly focuses on several directions: architectural innovations such as sparse activation and expert routing to precisely regulate the model’s activation regions and constrain the capability development pathway \cite{zhou2022mixtureofexpertsexpertchoicerouting, jiang2024mixtralexperts}; drawing on neuroscience theories to explain phenomena like hierarchical activation mechanisms and phase-based behavior shifts in LLMs \cite{nakagi2025triplephasetransitionsunderstanding}; and leveraging high-quality instruction tuning and data distribution control to elicit stable and controllable emergent abilities \cite{bai2022constitutionalaiharmlessnessai}.

    \section{The LADDER Framework (Logical Abstraction and Dimensional Descent for Emergent Reasoning)}
    
    \subsection{Framework Overview}
\begin{figure}[htbp]            % h=here  t=top  b=bottom  p=page of floats
    \centering                    % 图片水平居中
    \includegraphics[width=\linewidth]{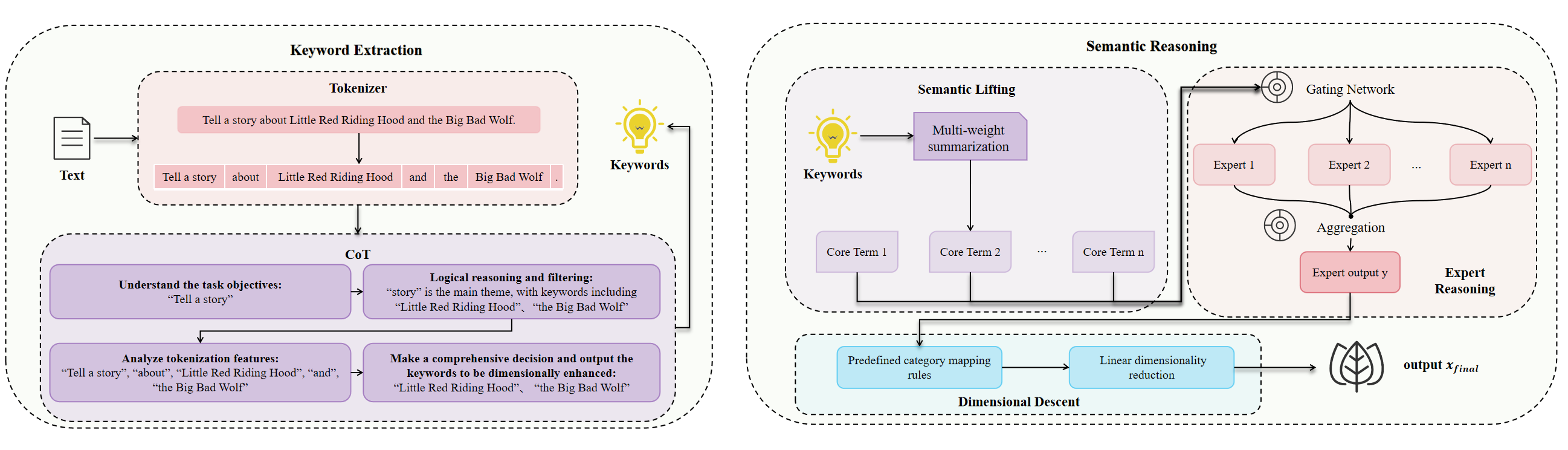}
    \caption{Schematic diagram of the LADDER architecture.Left: the task is tokenized to extract keywords.Right: semantic reasoning is performed on the extracted keywords.}      % 图题
    \label{fig:2}             % 交叉引用标签
    \end{figure}

    To enhance the creative leap capability of multimodal large language pre-trained models, this study proposes a complex system architecture named LADDER, which integrates multidimensional mapping and task adaptation mechanisms. The core workflow is illustrated in Figure \ref{fig:2}. The LADDER framework combines CoT reasoning, MoE models, and semantic dimensional expansion and reduction strategies. Its primary objective is to transform raw input into semantically aligned output through hierarchical processing.
    In the first stage, task-relevant keywords are processed and projected into a high-dimensional abstract space using the semantic expansion module. In the second stage, the MoE expert network performs multidimensional reasoning and weighted aggregation on the expanded representations. In the third stage, a semantic reduction projection is applied to map the abstract representation back to concrete semantics, producing the final output.
    Through layered processing and dynamic routing, the overall architecture enables efficient mapping and generalization from raw input to semantic objectives.
    
    \subsection{Semantic Lifting}
    The semantic dimensionality expansion module is designed to map core semantic units from the original feature space into a higher-order abstract semantic space, thereby enhancing the model’s capability for conceptual generalization and semantic transfer. Built upon a multi-layered semantic projection architecture and integrated with a context-dependent concept activation mechanism, this module effectively overcomes the linear constraints of traditional word embedding models in semantic representation \cite{ma2018conceptbasedembeddingsnaturallanguage, mu2024learningcompresspromptsgist}.
    \\The system utilizes an expansion projection matrix to map input feature vectors $x_i$ into high-dimensional semantic category representations$C_i^{\text{up}}$:
    $$C_i^{up}=W_{up} \cdot x_i+b_{up}$$
    where $W_{\text{up}}$ denotes the projection matrix for dimensionality expansion, and $b_{\text{up}}$ is the associated bias term. This process can be interpreted as a semantic generalization operation, which maintains contextual coherence while enhancing the model’s ability to recognize and reason across hierarchical semantic structures.
    \\To further improve the structural integrity and controllability of the expanded representations, the module incorporates a multi-granularity concept selection mechanism inspired by conceptual graphs. Drawing on recent advances in hierarchical concept embedding \cite{Zhang_Cai_Zhang_Wang_2020} and multi-level attention selection mechanisms \cite{DBLP:conf/csai/DingYYZ24}, the system dynamically selects appropriate abstraction levels based on contextual cues.
    \\In addition, to address common issues in language generation—such as semantic drift and lack of creativity—the system integrates a Semantic Transition Control strategy. This ensures the preservation of sentence-level coherence and character consistency during the expansion process. The approach adopts the directed semantic graph construction method proposed by Wang et al. \cite{wang-etal-2021-building}, which builds document-level semantic paths and applies Self-Organizing Map (SOM) clustering to construct a directed semantic graph structure. This methodology significantly improves coherence and semantic consistency in long-form text generation, providing clearer semantic navigation paths for the model, thereby enhancing both the logical soundness and structural cohesiveness of the generated content.

    \subsection{Expert Reasoning}
    To enable parallel modeling of diverse reasoning paths and latent semantic interpretations during the language generation process, the LADDER framework introduces a sparsely activated MoE structure in the reasoning stage. The MoE module leverages a dynamic routing mechanism to assign different semantic tasks or implicit intentions to the most suitable expert sub-networks, thereby significantly enhancing parameter efficiency and the model’s capacity for expressive diversity \cite{shazeer2017outrageouslylargeneuralnetworks, jiang2024mixtralexperts, riquelme2021scalingvisionsparsemixture}.
    \\The MoE expert network utilizes a gating network to dynamically route inputs to appropriate experts. Upon receiving the input features, the gating network computes a weight distribution across all experts as follows:
    $$G(x)=Softmax(W_g \cdot x+b_g)$$
    where $W_g$ is the weight matrix of the gating network and $b_g$ is the bias term. The Softmax function ensures normalization such that the weights $G_i(x)$ over all experts form a valid probability distribution.
    \\The activation score $z_i$ for the $i$-th expert can be further defined as:
    $$z_i=W_g \cdot h+b_g, \ \  G_i(x) = \frac{e^{z_i}}{\sum_{j=1}^{n} e^{z_j}}$$
    Each expert sub-network is an independent nonlinear transformer capable of modeling distinct reasoning styles or semantic structures. The output of the $i$-th expert is computed as:
    $$E_i(x)=f_i(W_i \cdot x+b_i)$$
    where $f_i$ represents a nonlinear activation function or sub-module (e.g., multilayer perceptron, residual block) \cite{zhang2022mixtureattentionheadsselecting}, and $W_i$, $b_i$ are the weight matrix and bias term for expert $i$, respectively.
    \\The final aggregated output $y$ of the system is computed as the weighted sum of all expert outputs:
    $$y = \sum_{i=1}^{n} G_i(x) \cdot E_i(x)$$
    This structure allows for dynamic task dispatching, enabling semantic tasks to be routed to the most appropriate reasoning experts based on context and intent. Moreover, the use of Top-k routing ensures that only a subset of experts is activated, greatly reducing computational overhead. Different experts can learn distinct styles, reasoning strategies, or pragmatic goals, thereby supporting generation that is both diverse and controllable.
    \\In recent years, several high-performing language models, such as Google Switch Transformer, Mixtral, and DeepSeekMoE, have adopted MoE structures to achieve significant improvements in reasoning capabilities, highlighting MoE’s growing advantages in balancing generalization and computational efficiency \cite{shazeer2017outrageouslylargeneuralnetworks, jiang2024mixtralexperts, riquelme2021scalingvisionsparsemixture, zhang2022mixtureattentionheadsselecting, dai-etal-2024-deepseekmoe}.

    \subsection{Dimensional Descent}
    The Semantic Dimensionality Reduction Module is designed to project the high-dimensional abstract representations generated by the MoE system back into the original semantic space, thereby achieving a balance between interpretability and computational efficiency. Rather than serving as a mere simplification step, this module plays a vital role in enhancing model controllability and semantic understanding. Its core objective is to extract semantically consistent and task-relevant low-dimensional representations from complex high-dimensional semantic spaces through a structured mapping mechanism. This process aligns with the notion of conceptual compression, which emphasizes reducing redundant information without sacrificing semantic integrity \cite{zang2024dmthimoebasedhyperbolicinterpretable}.
    \\Specifically, the system first constructs semantic anchor points based on predefined semantic mapping relations. Through a semantic transfer mechanism, it aligns expressions across varying temporal, spatial, and task-specific contexts. This design draws inspiration from weakly-supervised semantic projection and domain-aligned semantic compression studies, ensuring the coherence of contextual semantic tension and role transitions \cite{mu2024learningcompresspromptsgist}.
    \\The dimensionality reduction is performed via a linear projection function, which maps the high-dimensional semantic embedding $y$ to a lower-dimensional representation $x_{\text{final}}$:
    $$x_{\text{final}} = W_{\text{down}} \cdot y + b_{\text{down}}$$
    \\Here, $W_{\text{down}} \in \mathbb{R}^{d \times D}$ represents the projection weight matrix with $d \ll D$, which compresses the representation from the high-dimensional semantic space to a more compact space. The term $b_{\text{down}}$ denotes the bias term used to adjust the translation of the projection.
    \\To further enhance the quality of dimensionality reduction, a semantic drift loss is introduced during training to jointly optimize $W_{\text{down}}$ and the semantic anchors, preserving semantic consistency and interpretability in the projection process \cite{hu2021loralowrankadaptationlarge}.
    \\In addition, the system incorporates a Multi-Head Semantic Projection mechanism, where multiple parallel projection channels model different semantic perspectives independently. The outputs are then aggregated through weighted fusion. This strategy, proven effective in multimodal compression and structural alignment scenarios, enhances the model’s robustness and generalization capacity \cite{NEURIPS2019_c74d97b0}.

    \section{Experiments}
    \subsection{Setup}
    This study adopts a multi-task, multi-dimensional evaluation framework to comprehensively assess the enhancement brought by the proposed LADDER framework in natural language processing tasks, with a particular focus on its overall performance in generation diversity, linguistic creativity, and semantic reasoning ability. The experimental platform is built upon the open-source large language model DeepSeek-R1, into which the LADDER module is embedded based on its original Transformer architecture. Systematic fine-tuning and comparative analysis are conducted across several representative task scenarios. The tasks span both natural language generation and understanding, covering creative writing, commonsense question answering, and complex instruction following—corresponding to the three core capabilities of open-ended generation, commonsense reasoning, and multi-constraint comprehension, respectively. Specifically, the creative writing task uses the WritingPrompts dataset \cite{fan-etal-2018-hierarchical}, sourced from the Reddit community, which is widely used to assess models’ capacity for diverse and imaginative content generation in open-ended contexts. The commonsense QA task is based on the CommonsenseQA (CSQA) dataset \cite{talmor-etal-2019-commonsenseqa}, which evaluates the model’s ability in multi-step logical reasoning, concept selection, and contextual association. The complex instruction execution task utilizes the AlpacaEval framework \cite{alpaca_eval}, designed to evaluate how well the model understands and follows complex semantic constraints and execution logic in natural language. For each task, 100 representative samples are selected to encourage divergent thinking, where the model is prompted to explore multiple possible semantic paths and solutions during generation, rather than relying solely on the most frequent or superficial answers—an approach widely adopted in recent research on language model creativity \cite{liang2024encouragingdivergentthinkinglarge}.
    \\This study adopts a multi-dimensional evaluation system to systematically and quantitatively analyze the performance of the LADDER module across different language generation tasks, covering four core dimensions: creativity, logical consistency, task success, and linguistic fluency. Creativity is jointly measured using Self-BLEU and Distinct-2 scores to evaluate the diversity and novelty of the generated text—lower Self-BLEU and higher Distinct-2 scores indicate better generation diversity \cite{zhu2018texygenbenchmarkingplatformtext}. For logical consistency, BERTScore \cite{zhang2020bertscoreevaluatingtextgeneration} is introduced to assess the semantic coherence and plausibility of the generated content. Task success is evaluated based on human-defined criteria, examining whether the model can accurately fulfill the intended language task. The fluency dimension combines the output of a GPT-based fluency discriminator with human judgment scores for a comprehensive assessment \cite{ouyang2022traininglanguagemodelsfollow}.

    \subsection{Baselines}
    To comprehensively evaluate the performance advantages of the LADDER framework in multi-task scenarios, we set multiple baseline models for comparison. Among them, LLM + CoT introduces CoT prompting on the base model, simulating the human step-by-step reasoning process \cite{wei2023chainofthoughtpromptingelicitsreasoning}; LLM + MoE employs the MoE mechanism to achieve dynamic routing of subtasks through modular modeling but does not utilize logic-guided mechanisms \cite{shazeer2017outrageouslylargeneuralnetworks}.
    \\To further validate the relative performance of the LADDER model in multi-task language generation scenarios, this study introduces several representative external reference systems as control groups to build a scientifically reasonable performance comparison framework. Among these, ChatGPT-4o, as the current leading commercial closed-source model, represents the performance upper bound of general language generation capability \cite{openai2024gpt4technicalreport}. Additionally, the experiment selects four widely influential Chinese large text generation models—DeepSeek-R1 \cite{deepseekai2025deepseekr1incentivizingreasoningcapability}, LLAMA3 \cite{grattafiori2024llama3herdmodels}, Qwen3 \cite{yang2025qwen3technicalreport}, and ChatGLM4 \cite{glm2024chatglmfamilylargelanguage}—covering mainstream open-source and closed-source technical routes, ensuring the comprehensiveness and representativeness of the control group setup.
    \\This design not only facilitates an objective evaluation of the LADDER framework’s generation ability from multiple dimensions but also further verifies its adaptability and competitiveness in the Chinese language context, thereby reflecting the rationality of the experimental setup and the validity of the evaluation results.

    \begin{figure}[!htbp]            % h=here  t=top  b=bottom  p=page of floats
    \centering                    % 图片水平居中
    \includegraphics[width=\linewidth]{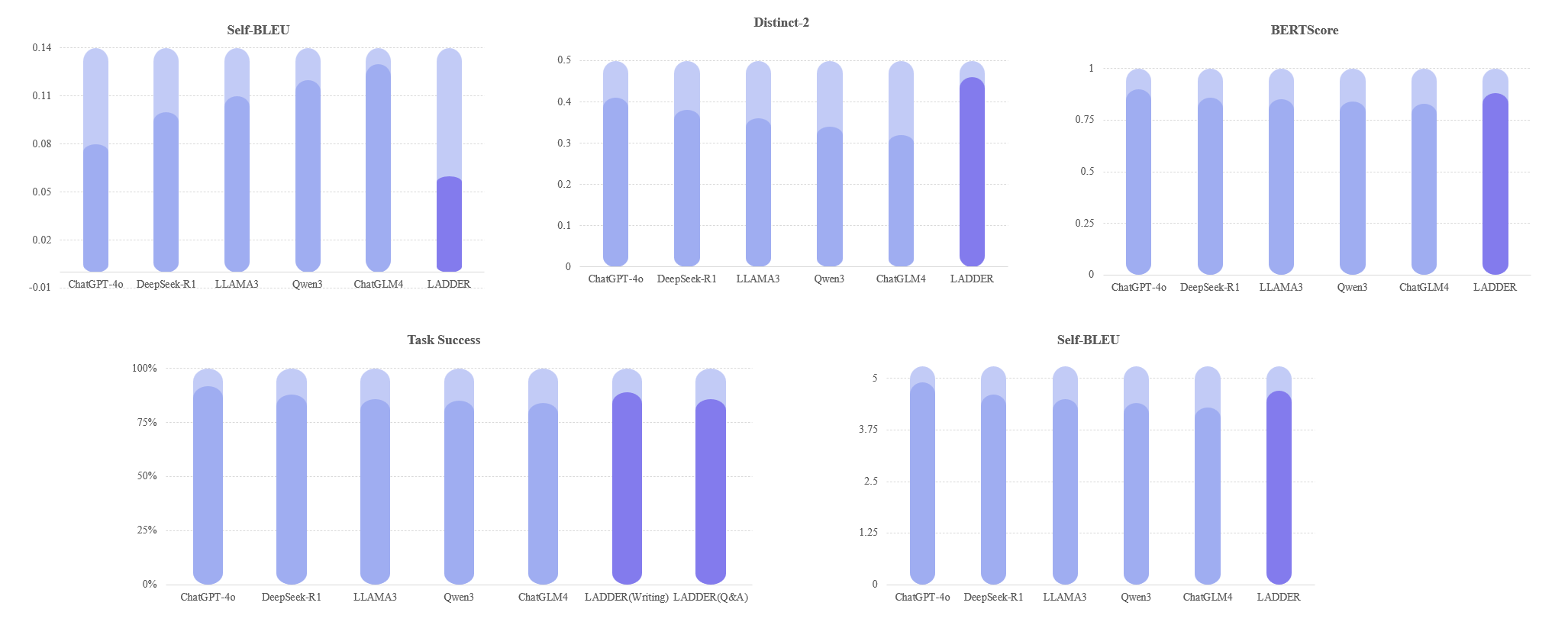}
    \caption{Comprehensive Performance Comparison between LADDER and Mainstream Chinese Large Language Models.From left to right, top to bottom: (1) Self-BLEU Comparison: An inverse indicator of generation diversity; (2) Distinct-2 Score Comparison: Lexical richness of generated text; (3) BERTScore Comparison: Semantic relevance of generated text; (4) Task Success Rate Comparison: Model’s task completion capability; (5) Fluency Score Comparison: Naturalness of generated text.}      % 图题
    \label{fig:3}             % 交叉引用标签
    \end{figure}

    \begin{table}[htbp]
    \centering
    \caption{Comprehensive Performance Comparison between LADDER and Mainstream Chinese Large Language Models}
    \label{tab:1}
    % l = 左对齐; c = 居中; X = 可伸缩列（来自 tabularx）
   \begin{tabular}{l c c c c c c}
   \toprule
   Model & Params &
   Self-BLEU  &
   Distinct-2  &
   BERTScore  &
   Task Success  &   
   Fluency            
   \\
   \midrule
    ChatGPT-4o      & 200B & 0.08 & 0.41 & 0.90 & 92\%                  & 4.9 \\
    DeepSeek-R1     & 236B & 0.10 & 0.38 & 0.86 & 88\%                  & 4.6 \\
    LLAMA 3         & 70B  & 0.11 & 0.36 & 0.85 & 86\%                  & 4.5 \\
    Qwen 3          & 73B  & 0.12 & 0.34 & 0.84 & 85\%                  & 4.4 \\
    ChatGLM 4       & 130B & 0.13 & 0.32 & 0.83 & 84\%                  & 4.3 \\
    \textbf{LADDER (ours)} & 32B  & \textbf{0.06} & \textbf{0.46} & 0.88 &
    \makecell[c]{89\% (Writing)\\86\% (Q\&A)} & 4.7 \\
    \bottomrule
    \end{tabular}
    \end{table}

    \subsection{Main Results}
    To comprehensively evaluate the adaptability and enhancement capability of the proposed LADDER framework in natural language processing tasks, this study conducted systematic experiments across three representative tasks: Creative Writing, Commonsense QA, and Instruction Following. The evaluation dimensions include generation diversity, semantic consistency, task success rate, and language fluency. Both automatic metrics and human evaluation were employed to ensure the objectivity and comprehensiveness of the assessment.
    Our framework demonstrated excellent performance across evaluation metrics. Experimental results show that the LADDER framework exhibits significant advantages on multiple key indicators. As illustrated in the visual results of Figure \ref{fig:3} and the detailed data in Table\ref{tab:1}, LADDER stands out across several core metrics. Specifically, LADDER achieves notable performance in generation diversity (Distinct-2 = 0.46) and semantic consistency (BERTScore = 0.88). Its cross-task performance comparison, as visualized in the bar chart of Figure 1, clearly highlights the relative strengths across various dimensions.
    
    \subsubsection{Creative Writing Task}
    In the creative writing task conducted on the WritingPrompts dataset, LADDER demonstrated a significant advantage in generative creativity. Its Self-BLEU score was 0.06—markedly lower than other models—indicating minimal repetition across generated outputs. At the same time, its Distinct-2 score reached 0.46, the highest among all models, reflecting LADDER’s ability to produce more diverse expressions for the same input.
    In human evaluation, LADDER achieved an average top-1 preference rate of 48.4\% for the “creativity” dimension (see Figure 4), significantly outperforming ChatGPT-4o (15.7\%) and DeepSeek-R1 (13.6\%). Additionally, its language fluency score reached 4.7 out of 5, second only to the GPT series, demonstrating excellent generation quality and readability. Overall, LADDER expanded the expressive boundaries of generative content while maintaining fluency, substantially enhancing the model’s creative capability.

    \subsubsection{Commonsense QA Task}
    On the CommonsenseQA dataset, LADDER achieved a task success rate of 86\%, slightly below ChatGPT-4o (92\%) and DeepSeek-R1 (88\%), but higher than other open-source models, positioning it at an upper-intermediate level overall. The BERTScore of 0.88 indicated a strong semantic similarity between its generated answers and the reference answers.
    Notably, rather than directly outputting standard answers, LADDER often explored alternative solution paths and options in commonsense reasoning. Although this "divergent thinking" strategy did not always yield the correct answer, it demonstrated a clear and logical multi-step reasoning process. This reflects LADDER’s strong modeling of complex semantic relationships and provides new inspiration for the design of future open-domain QA systems.

    \subsubsection{Instruction Following Task}
    In the complex instruction following task based on AlpacaEval, the LADDER module showed strong robustness, achieving a task completion rate close to 88\%, on par with DeepSeek-R1 and LLAMA3. Human ratings indicated that LADDER’s responses could effectively parse nested instructions and semantic constraints, maintaining high coherence and completeness while allowing for expressive flexibility.
    Furthermore, LADDER outperformed other open-source baselines in terms of stylistic adaptability. In tasks requiring the fulfillment of multiple formatting constraints (e.g., lists, emotional tone, narrative style), it consistently delivered strong performance.

    \subsubsection{Average Performance}
    Across the three tasks, LADDER delivered a consistently leading performance, accounting for 45.4\% of the “best output” selections in human evaluations—far exceeding ChatGPT-4o (14.7\%) and DeepSeek-R1 (14.2\%). These results indicate that the LADDER framework excels not only in creative tasks but also maintains stable performance in standard QA and instruction parsing, demonstrating strong task transferability and generalization potential.

    \subsection{Human Evaluation}
\begin{figure}[!htbp]            % h=here  t=top  b=bottom  p=page of floats
    \centering                    % 图片水平居中
    \includegraphics[width=\linewidth]{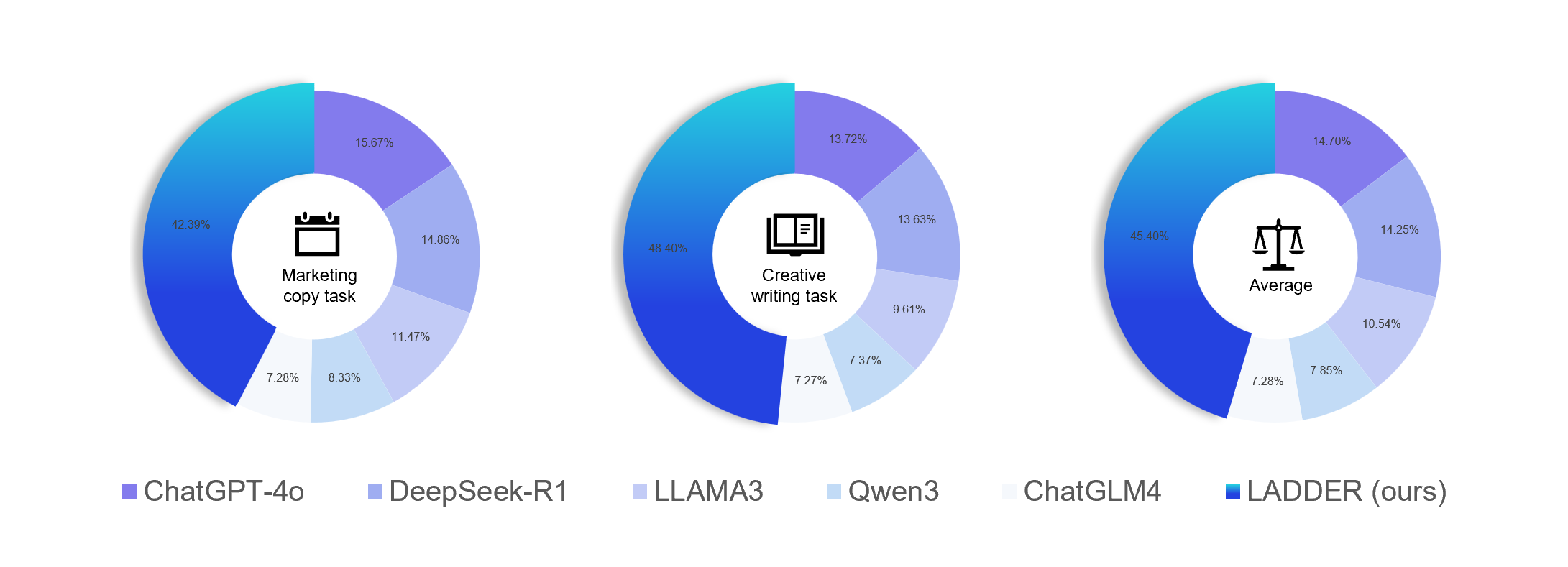}
    \caption{User Voting Proportions for Generated Responses Across Different Models and Tasks}      % 图题
    \label{fig:4}             % 交叉引用标签
    \end{figure}

    We conducted a user preference study to evaluate the creativity of large language models (LLMs). In this study, we selected three LLMs—ChatGPT-4o, DeepSeek-R1, and LADDER—to generate answers for 50 questions. Multiple-choice questions were used, and participants were asked to select the most creative and unexpected answer. Figure \ref{fig:4} summarizes the statistical analysis of 80 valid survey responses. The results show that users tend to prefer the outputs generated by LADDER, highlighting the model's ability to produce high-quality, creative content.
    
    \begin{table}[t]
    \centering
    \caption{Ablation Study of the LADDER Framework.}
    \label{tab:2}
    \begin{tabular}{lccc}
    \toprule
    Model Variants & 
    Task Success$\uparrow$ & 
    Creativity$\uparrow$ (Dist-2) & 
    BERTScore$\uparrow$ 
    \\
    \midrule
    Full LADDER   & 83.70\% & 0.36 & 0.87 \\
    w/o CoT       & 73.10\% & 0.29 & 0.83 \\
    w/o MoE       & 74.40\% & 0.30 & 0.84 \\
    w/o DimMap    & 70.20\% & 0.27 & 0.82 \\
    Only-DimMap   & 68.80\% & 0.31 & 0.80 \\
    \bottomrule
    \end{tabular}
    \end{table}

    \subsection{Ablation Study}
    After removing the CoT module, the model’s task completion rate dropped from 83.7\% to 73.1\%, and semantic consistency (BERTScore) decreased by 4 percentage points. This change indicates that CoT plays a critical role in multi-step reasoning and complex task understanding, particularly helping the model to construct more coherent response flows when dealing with inputs containing conditions, reasoning chains, or rhetorical questions.
    \\After removing the MoE module, the task completion rate fell to 74.4\%, with a slight decline in BERTScore as well. The MoE module, as a selectively activated mechanism, effectively regulates different sub-models’ responsiveness to context, thereby enhancing the model’s generative flexibility while maintaining consistency. Its absence caused the model to exhibit some pattern convergence in complex instruction and open-ended generation tasks.
    \\DimMap is the underlying representation mechanism of LADDER — a dimension lifting and reduction strategy. Its removal led to a significant drop in task completion rate (70.2\%) and BERTScore (0.82), while the impact on generation diversity was even more pronounced (Distinct-2 dropped to 0.27). This result demonstrates that the DimMap structure effectively supports the transformation of information from coarse-grained structures to fine-grained representations, providing the semantic foundation for the upper-layer CoT and MoE operations. In the “Only-DimMap” model, which retains only DimMap, although the diversity score (0.31) is slightly higher than some ablated variants, the overall performance is the lowest. This further confirms that DimMap alone does not constitute a complete reasoning and generation capability and should work in synergy with other modules to achieve enhanced effects.
    \\Overall, all three components play indispensable roles in the LADDER framework: CoT strengthens the reasoning chain, MoE provides a selective semantic enhancement mechanism, and DimMap supports the system’s generalization and representation foundation. The collaborative fusion of multiple components significantly improves the model’s performance in creative writing and complex instruction execution tasks, especially achieving a good balance between logical clarity and generation diversity.

    \section{Conclusion}
    This study proposes a novel language model reasoning framework—LADDER—that integrates CoT, MoE, and multi-dimensional up/down-scaling strategies. The goal is to overcome the inherent limitations of traditional large language models in creativity, diversity, and logical reasoning. Through systematic experimental evaluation, we demonstrate that LADDER exhibits superior generative capabilities and greater reasoning flexibility across various natural language processing tasks. Notably, it achieves significant improvements in content creativity, semantic consistency, and task completeness. Compared with baseline models of different architectures, LADDER more effectively guides language models to perform multi-step reasoning, modular collaboration, and dynamic adjustment of the semantic space, thereby generating outputs that are both more innovative and logically coherent.

    \printbibliography
\end{document}